\documentclass[runningheads]{llncs}
\usepackage{graphicx}

\usepackage{tikz}
\usepackage{comment}
\usepackage{amsmath,amssymb} 
\usepackage{color}

\usepackage[accsupp]{axessibility}  

\usepackage{hhline}
\usepackage{multirow}
\usepackage{pifont}
\usepackage{bbm}
\usepackage{cite}
\usepackage{algorithm}
\usepackage{algpseudocode}
\usepackage{svg}
\usepackage{xcolor}
\usepackage{booktabs}
\usepackage{colortbl}
\usepackage{appendix}
\usepackage[normalem]{ulem}
\usepackage{graphics}
\usepackage[colorlinks=true]{hyperref}
\usepackage[capitalise]{cleveref}
\usepackage{caption}
\usepackage{subcaption}
\usepackage{kotex}
\usepackage{todonotes}

\newcommand{\cmark}{\ding{51}}
\newcommand{\xmark}{\ding{55}}
\DeclareMathOperator*{\E}{\mathbb{E}}

\definecolor{Gray}{gray}{0.85}
\newcommand{\ie}{\textit{i.e.}}
\newcommand{\eg}{\textit{e.g.}}

\begin{document}

\pagestyle{headings}
\mainmatter

\title{DLCFT: Deep Linear Continual Fine-Tuning\\for General Incremental Learning}


\titlerunning{Deep Linear Continual Fine-Tuning for General Incremental Learning}
\author{Hyounguk Shon\inst{1}\thanks{Work done during an internship at LG AI Research.} \and
Janghyeon Lee\inst{2} \and
Seung Hwan Kim\inst{2}\index{Kim, Seung Hwan} \and
Junmo Kim\inst{1}}
\authorrunning{H. Shon et al.}

\institute{Korea Advanced Institute of Science and Technology \and LG AI Research
\\ \email{\{hyounguk.shon, junmo.kim\}@kaist.ac.kr}
\\\email{\{janghyeon.lee, sh.kim\}@lgresearch.ai}
}

\maketitle

\begin{abstract}
Pre-trained representation is one of the key elements in the success of modern deep learning. However, existing works on continual learning methods have mostly focused on learning models incrementally from scratch.
In this paper, we explore an alternative framework to incremental learning where we \emph{continually fine-tune} the model from a pre-trained representation. Our method takes advantage of linearization technique of a pre-trained neural network for simple and effective continual learning. 
We show that this allows us to design a linear model where quadratic parameter regularization method is placed as the optimal continual learning policy, and at the same time enjoying the high performance of neural networks. We also show that the proposed algorithm enables parameter regularization methods to be applied to class-incremental problems.
Additionally, we provide a theoretical reason why the existing parameter-space regularization algorithms such as EWC underperform on neural networks trained with cross-entropy loss. We show that the proposed method can prevent forgetting while achieving high continual fine-tuning performance on image classification tasks. To show that our method can be applied to general continual learning settings, we evaluate our method in data-incremental, task-incremental, and class-incremental learning problems.

\end{abstract}

\section{Introduction}

The ability to incrementally accumulate knowledge from a sequence of datasets is a crucial functionality that modern AI systems require. It is well known that deep neural networks suffer from significant performance degradation when the learning is done sequentially. Such phenomena is referred as catastrophic forgetting (CF), which continual learning aims to address. 

Transfer learning is one of the key contributing elements in the recent success of deep learning across various applications from visual to linguistic tasks. When dealing with visual signals, neural nets are often pre-trained on large datasets such as ImageNet~\cite{ILSVRC15} before training on the target task, which often brings significant performance boost. 
Although fine-tuning from a pre-trained representation is a standard practice adopted in a lot of modern deep learning applications, many of the existing approaches to continual learning assume a scenario where one needs to begin training from a randomly initialized model. In this work, we explore a practical alternative continual learning framework based on pre-training of neural network, coined \emph{continual fine-tuning}.

Existing works on continual learning can be largely categorized into three groups: model regularization, data rehearsal, and parameter isolation~\cite{delange2021continual}. Regul\-arization-based methods aim to penalize the update either in function space \cite{LWF18, LwM19} or in parameter space~\cite{EWC2017, OSLA2018, XKFAC20, mas18, ucb20, vcl18, kurle2019continual, si17, IMM17}. Parameter isolation methods~\cite{mallya2018packnet, Yoon2020Scalable} update the architecture in order to isolate the knowledge learned from each task in order to prevent forgetting. Rehearsal-based approaches~\cite{icarl17, gdumb20, gem17, agem19} replay examples from previous tasks either by storing samples to an external memory buffer or learning a generative model. The rehearsal method has shown to be effective at regularization with a small extra cost.

Parameter regularization methods based on Fisher information matrix aim to represent the source task objective by second-order Taylor approximation, which is typically too expensive due to the quadratic memory cost of the Hessian matrix. Naturally, existing works have focused on efficient representation of the matrix via diagonal approximation\cite{EWC2017,R-EWC18} or Kronecker factorization\cite{OSLA2018,XKFAC20}. Our work is motivated by a question that has been relatively overlooked in prior works: ``Does better Hessian approximation improve continual learning?''

Although parameter regularization approaches are founded on a principled framework that computes the importance of the weights for each task, they have shown relatively underwhelming performance compared to rehearsal-based approaches. One of the main roadblocks of the regularization methods is to address the problems coming from their high parameter dimension and non-linearity, which has left continual learning a particularly challenging problem to tackle. In this work, we show that it is possible to bring a significant boost to regularization methods by a simple modification on the loss function and reparametrization of the model.
\begin{figure}[t]
\centering
\includegraphics[width=0.8\textwidth]{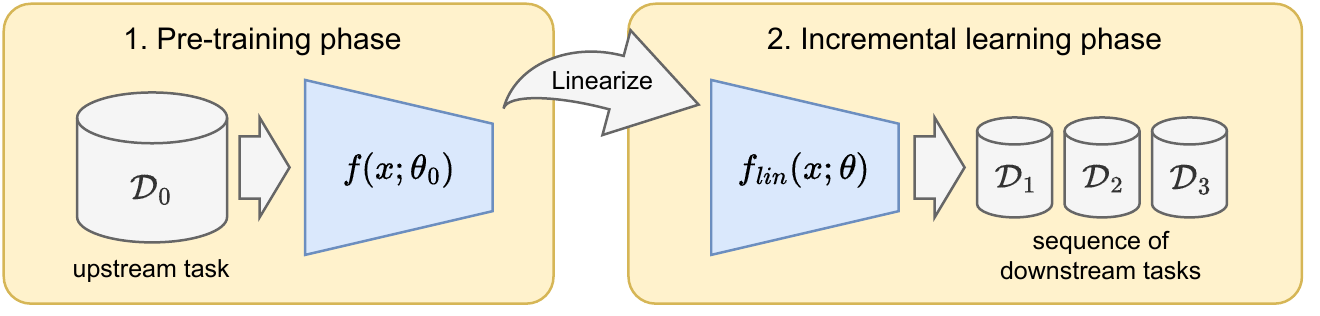}
\caption{\textbf{Proposed continual fine-tuning framework.} The model is first trained on a large dataset (e.g., ImageNet) to obtain general representation. At incremental learning phase, the model is continually fine-tuned and accumulates knowledge by learning from the sequentially arriving batches of dataset.}
\label{fig:linear-continual-learning}
\end{figure}
The remaining portion of the paper is organized as follows: In \cref{sec:prelimiaries}, we provide background on continual learning methods, second-order derivative of neural networks, and linearization of neural networks. In \cref{subsec:problems-of-parameter-regularization}, we provide reasoning on why existing parameter regularization methods underperform in mitigating catastrophic forgetting. In \cref{subsec:dlcft}, we describe the proposed method, Deep Linear Continual Fine-tuning (DLCFT).  In \cref{subsec:dlcft-class-il}, we elaborate on how the proposed regularization method can be applied to the class-incremental problem where existing parameter regularization methods have been unable to be applied. In \cref{sec:experiments}, we show the evaluation and analysis of the proposed method on data-/task-/class-incremental learning problems. Finally, in \cref{sec:conclusion}, we conclude the paper.

The main contributions of this work are summarized as follows:
\begin{itemize}
\item Instead of incrementally training a neural network from scratch, we propose an alternative approach to continual learning that leverages pre-trained representations, coined \emph{continual fine-tuning}. We show that our approach introduces a novel method for simple and practical continual learning in deep learning.
\item To continually adapt to a sequence of downstream tasks, we utilize pre-trained neural network through decomposing the model into nonlinear and linear components by linearization. We propose a learning algorithm that combines  linearization and mean squared error loss that significantly boosts the effectiveness of quadratic weight regularization methods. Further, we provide justification on why linearization is the key component for a principled approach to optimal continual learning. 
\item Our method can be universally applied to various continual learning scenarios where new batches of data, task, or class are observed sequentially. Although data-incremental learning is an important open challenge to practical deep learning, relatively little attention has been given from the community. Notably, we show that our method can effectively learn in data-incremental scenario where batches of new data samples are observed sequentially. Additionally, we demonstrate our method in task-incremental and class-incremental learning scenarios with a small memory buffer.
\end{itemize}

\section{Preliminaries}
\label{sec:prelimiaries}

\subsection{Continual learning}

Existing approaches to continual learning can be largely grouped into three families: regularization-based, rehearsal-based, and parameter isolation methods. Regularization method incorporates additional training objectives to prevent model from changing too much during training. One can regularize the outputs of the model to mitigate forgetting. Learning without forgetting (LwF)~\cite{LWF18} is one of the early works that utilizes target task data as a surrogate for the source task samples. This method shows reasonably good performance when the source task domains and target task domains are similar, however 
 is less effective when the task domains are dissimilar.

Parameter regularization methods aim to estimate the importance of parameters and use that as the prior for the parameters during training of the subsequent tasks. This was first adopted by elastic weight consolidation (EWC)\cite{EWC2017} where the authors proposed to use the diagonal entries of Fisher information matrix. Online-structured Laplace approximation (OSLA)\cite{OSLA2018} used Kronecker-factored approximate curvature (K-FAC) \cite{KFAC2015} to incorporate off-diagonals of the Fisher information matrix. With neural networks having intra-batch dependency due to batch normalization, Extended K-FAC (XK-FAC)~\cite{XKFAC20} generalized the method to take intra-batch dependency into account. 

Rehearsal-based methods maintain a buffer that stores a small number of samples, and replay the examples in the training of the subsequent tasks. One of the pioneering methods is iCaRL\cite{icarl17} which keeps track of per-class samples for class-incremental learning problems. Some other line of works such as GEM\cite{gem17} and A-GEM~\cite{agem19} performs constrained optimization by projecting the gradients so that they do not interfere with the previous tasks. DER~\cite{buzzega2020dark} proposed to distill from the logits that are sampled and stored throughout the training. Parameter isolation methods \cite{expertgate17, mallya2018packnet, rusu2016progressive, serra2018overcoming} are based on dynamically allocating a set of parameters for each task so that the training does not interfere with each other.

\subsection{Second-order derivatives of neural network}

Computing and storing the Hessian matrix of neural networks is a difficult challenge due to their high parameter dimension. Hessian matrix of a probabilistic model can be approximated using Fisher information matrix (FIM)~\cite{pascanu2013revisiting}. FIM can be interpreted as the second derivative of the KL-divergence between the model and target distribution, and can be efficiently estimated through Monte Carlo method.

Because the memory cost of the full Hessian matrix is quadratic to the size of the parameters, approximation or factorization technique is necessary to handle the matrix. One popular method is diagonal approximation of the Hessian, which neglects the influences of the off-digonal components. Another approach that can consider off-diagonal influences is block-diagonal approximation, which considers the correlations among intra-layer parameters. K-FAC~\cite{KFAC2015, kfc16} proposed to approximately factorize the block diagonals of a neural network's Fisher information matrix into a Kronecker product of covariance matrices: 
\begin{align}\begin{split}
    \label{eq:k-fac}
    F^{ll} & = \E_{y\sim p(y|x)}\left[\nabla_\theta \log{p(y|x)} \nabla_\theta \log{p(y|x)}^\top\right] \\
    & = \E[(g \otimes a) (g \otimes a)^\top] = \E[(g g^\top) \otimes (a a^\top)] \\
    & \approx \E[g g^\top] \otimes \E[a a^\top] ,
\end{split}\end{align}
where $F^{ll}$ is the FIM diagonal block of the $l$-th layer, $g$ is the pre-activation gradient, $a$ is the input activation, and $\otimes$ indicates Kronecker product. K-FAC represents the curvature matrix in the form of a product between two factors, thereby reducing the memory cost from $\mathcal{O}(M^2 \times N^2)$ to $\mathcal{O}(M^2 + N^2)$, where $M$ and $N$ are the sizes of the input and output units of the layer, respectively.

Eigenvalue-corrected K-FAC~\cite{ekfac18} proposed to re-compute the diagonals in the eigenbasis obtained from K-FAC, thereby correcting the eigenvalues of the factorized FIM. Trace-restricted K-FAC (TK-FAC)~\cite{tkfac21} is a trace-exact variant which corrects the norm of the factorized FIM by further tracking the trace of FIM.

\subsection{Linearization of deep neural networks}

Linear approximation in deep learning is a widely used concept as a means of estimating the proximal behavior of a neural network. Popular explanation methods such as Guided Backprop~\cite{guidedbackprop14} and Grad-CAM~\cite{gradcam16} use the gradient with respect to the activations in order to estimate the network's the sensitivity against the input. Such techniques are based on the first-order derivatives of neural networks with respect to the input.

The first-order derivatives with respect to the parameters, on the other hand, have lead to interesting insights about the training of neural networks. Recent works regarding the training dynamics of deep neural networks \cite{ntk18, lee2019wide} have found that randomly initialized neural networks behave linearly throughout gradient descent training in the infinite-width regime. In \cite{ntk18}, it is shown that such model can be described by a specific kernel function, coined neural tangent kernel, defined by the first-order derivatives of the neural network. Moreover, recent works \cite{GaF20, LQF21} have empirically shown that the observation also apply to finite-sized neural networks when they are pre-trained, and showed the linearized network can be fine-tuned to achieve comparable performance to the nonlinear network.

\begin{figure}[t]
\centering
\includegraphics[width=0.65\textwidth]{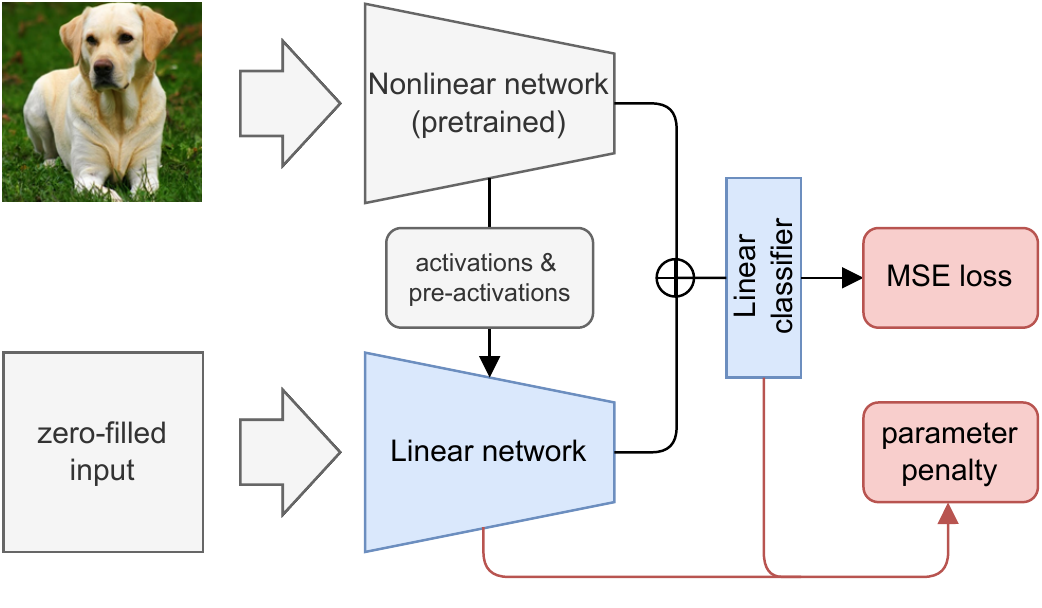}
\caption{\textbf{Architecture of our deep linear continual fine-tuning model.} \textcolor{blue}{Learnable modules} are colored in blue. On forward pass, the image and a zero-filled tensor are passed into the nonlinear network (fixed) and linear network (learnable), respectively. The nonlinear feature and the linear residual feature are computed concurrently, which are then summed and passed to the linear classification layer. Notice that the output logits are strictly linear with respect to all learnable parameters. At training, MSE loss is used instead of cross-entropy. During continual learning, model is regularized by quadratic parameter penalty loss to prevent forgetting.}
\label{fig:linear-network-architecture}
\end{figure}

\section{Continual Fine-tuning}

The goal of continual learning is to achieve the highest performance jointly on all tasks when the tasks arrive sequentially. Due to the sequential nature, the model is only allowed to observe a batch of data $\mathcal{D}_t$ at each task $t$. The performance of continual learning can be seen as upper bounded by multi-task learning, where the model trains jointly on all tasks,
\begin{align}
    \min_{\theta} \frac{1}{T} \sum_{t=1}^{T} \E_{(x,y) \in \mathcal{D}_t} \left[ \lambda_t \mathcal{L}(f(x;\theta, t),y) \right] ,
\end{align}
where $\theta$ is the vectorized parameter of the linearized model, and $\lambda_t$ controls the stability--plasticity between the tasks. To learn all tasks from sequentially arriving batches, parameter regularization methods aim to capture the source task objective as a quadratic function of parameters, \ie,
\begin{align}\begin{split}
\frac{1}{t} \sum_{i=1}^{t} \E_{(x,y) \in \mathcal{D}_i} \left[ \lambda_i \mathcal{L}(f(x;\theta,i),y) \right] 
&= C + \frac{1}{2} (\theta - \theta_t)^\top A (\theta - \theta_t) + \mathcal{O}(\theta^3)\\
&\simeq C + \frac{1}{2} (\theta - \theta_t)^\top A (\theta - \theta_t) ,
\end{split}\end{align}
where $A$ is the matrix whose each entry represents the importance of the corresponding parameter pair, $C$ is a constant, and $\theta_t$ is the model parameters after learning the $t$-th task.

Parameter-based continual learning methods aim to accurately estimate the importance matrix $A$ in order to closely approximate the multi-task learning objective. One popular approach is to use second-order Taylor approximation, where $A$ becomes the Hessian matrix of the source task objective function.

\subsection{Understanding the problems of quadratic parameter regularization}
\label{subsec:problems-of-parameter-regularization}

\begin{figure}[t]
\centering
    \begin{subfigure}{.4\linewidth}
        \centering
        \includegraphics[width=0.9\linewidth]{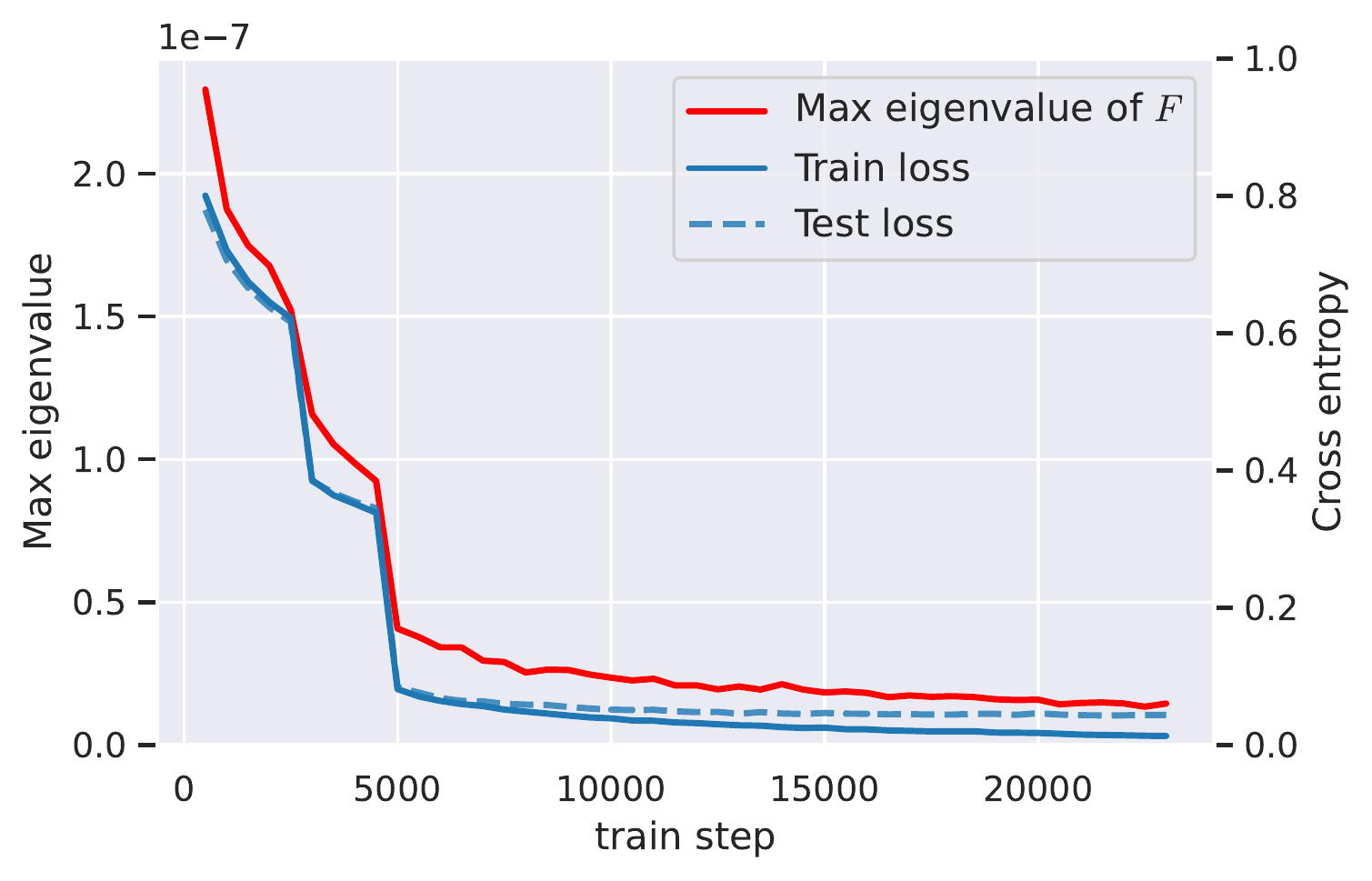}
        \caption{Curvature of SCE loss}
        \label{fig:diminishing-curvature-sce}
    \end{subfigure}
    \begin{subfigure}{.4\linewidth}
        \centering
        \includegraphics[width=0.9\linewidth]{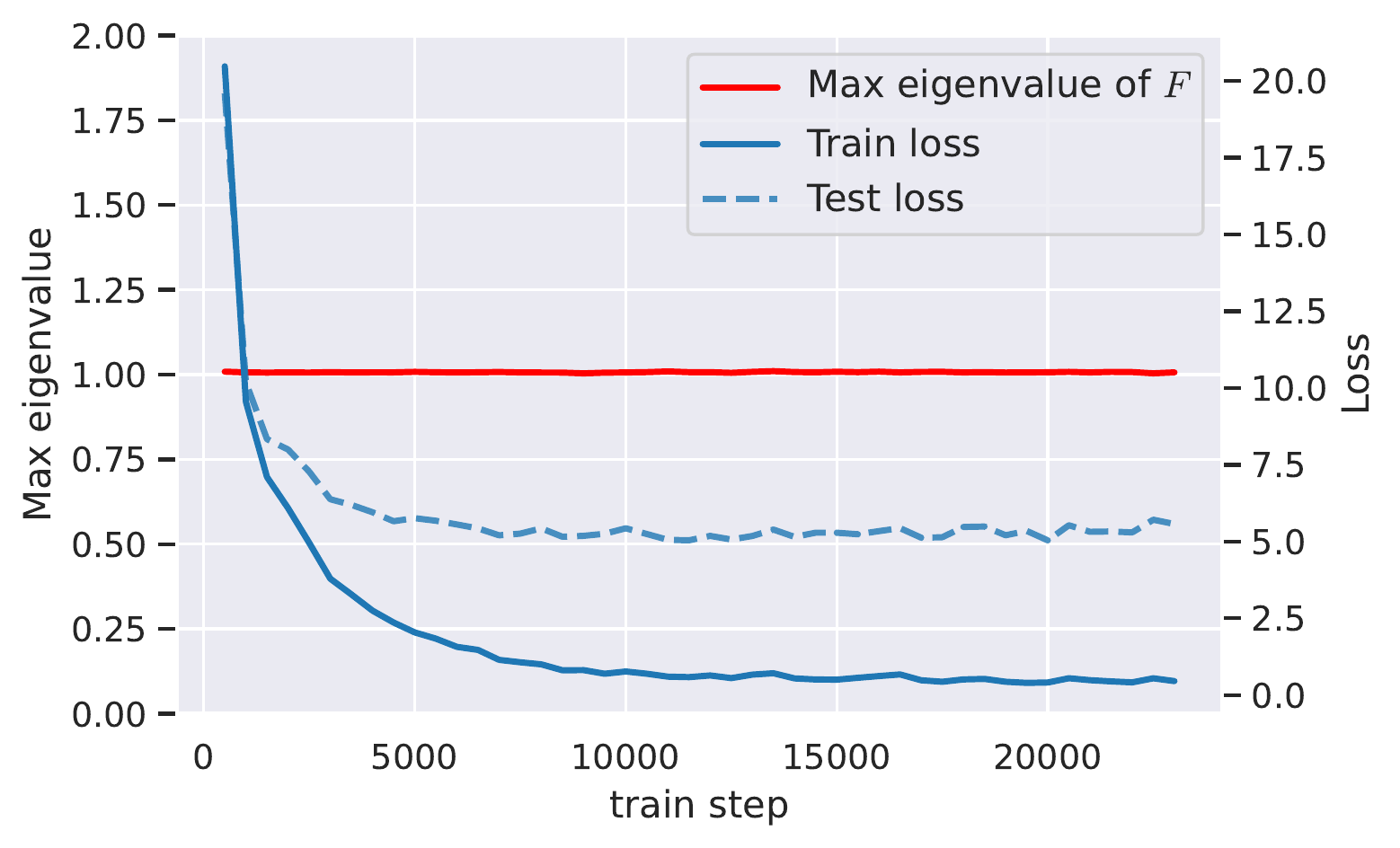}
        \caption{Curvature of MSE loss}
    \end{subfigure}
\caption{An MLP is trained on MNIST using softmax cross-entropy (SCE) and mean squared error (MSE) losses, and the maximum eigenvalue of the Fisher information matrix of each loss function is traced and plotted in red. The eigenvalues of SCE loss diminish to zero as the model fits the data, whereas that of MSE loss remains constant.}
\label{fig:diminishing-curvature}
\end{figure}

In this section, we provide why existing parameter regularizations show underwhelming performance in continual training of neural networks. We particularly look into two sources of problems which we call \emph{vanishing curvature} and \emph{higher-order error}. First, we show that softmax cross-entropy loss causes curvature to vanish to zero, thereby losing ability to represent parameter importance. Then, we show why this problem is amplified by nonlinearity of neural network.

First, we investigate the behavior of FIM of a model that predicts categorical distribution and trained using cross-entropy loss. The second order derivatives of a probabilistic model is often estimated through FIM, which is the covariance matrix of the gradient with respect to the log-likelihood.
\begin{equation}
    \label{eq:fisher-information}
    F_\theta = \E_{y\sim p_\theta(y|x)}{\left[ \nabla_\theta \log{p_\theta(y|x)} \nabla_\theta \log{p_\theta(y|x)}^\top \right]}
\end{equation}
It can be seen from \cref{eq:fisher-information} that as the model $p_\theta(y|x)$ perfectly fits the target label, the sampled $y$ becomes the target label with probability$\rightarrow$1 and  $\nabla_\theta \log{p_\theta(y|x)} \rightarrow 0$, therefore $F_\theta \rightarrow 0$. This indicates that FIM loses its ability to represent parameter importance at near zero-loss optima. See \cref{supp:sce-loss-hessian} in the supplementary for detailed proof. We name this \emph{vanishing curvature problem}. Additionally, notice that this behavior is caused by choosing to fit a categorical distribution, e.g., softmax layer.
\cref{fig:diminishing-curvature-sce} shows that the maximum eigenvalue of FIM continues to vanish to zero as the negative log-likelihood approaches to zero loss, even after the test loss has converged. As a result, the parameters are under-regularized.

Secondly, quadratic approximation assumes that the objective function is a quadratic function with respect to the parameters. However, the true loss function is non-quadratic due to the cross-entropy loss and the non-linearity of neural networks. This introduces higher-order error terms to dominate in the second-order approximation. 
Combined with the vanishing curvature behavior, this amplifies the error of the loss approximation as the parameters are under-damped and the model can drift off the trust region of the local loss approximation which causes catastrophic forgetting.

\subsection{Deep Linear Continual Fine-tuning (DLCFT)}
\label{subsec:dlcft}

To this end, we propose an alternative approach based on continual fine-tuning framework. Our change to the model is two-fold; To tackle the vanishing curvature problem, we replace the cross-entropy loss with MSE loss function. To resolve the non-convexity problem, we choose to \emph{approximate the model} such that it has simple linear structure. The combination of changes to the model allows parameter-based regularization to be the optimal continual learning policy.

Firstly, to work around the vanishing curvature problem, we replace softmax cross-entropy loss for MSE loss which is a non-saturating, quadratic loss function. 
\begin{align}\begin{split}
    \ell(f(x;\theta,t),y) = \frac{1}{2} ||\alpha \phi(y) - f(x;\theta,t)||^2
\end{split}\end{align}
Here, $\phi(\cdot)$ indicates the one-hot representation, and $\alpha$ is a positive scaling constant which is fixed throughout all tasks. 
We followed \cite{hui2020evaluation, LQF21} and set $\alpha=15$. 

Secondly, to tackle the higher-order error problem, we linearize the neural network. We apply first-order Taylor approximation of a pre-trained neural network \cite{GaF20, LQF21} which decomposes the feature extractor into a frozen non-linear network and a trainable linear network. Instead of fine-tuning the full non-linear neural network, we train the linearized neural network, \ie,
\begin{align}
    g_{lin}(x; \psi) & = g(x; \psi_0) + D_\psi g(x;\psi_0) \cdot \psi ,
\end{align}
where $D_\psi g(x;\psi_0)$ is the Jacobian of the network evaluated at the pre-trained point $\psi_0$. Then we learn a linear classification layer using the linearized feature, 
\begin{align}
    f(x; \theta,t) & = w_t \cdot g_{lin}(x; \psi) + b_t ,
\end{align}
where $g(x; \psi_0)$ is the pretrained nonlinear feature extraction network and $\theta = \{ \psi, w_t, b_t \}$ corresponds to the parameters of the linearized model. For data-incremental learning setup, we train the linearized feature extraction network and a single linear classifier. We regularize both the feature extractor and the linear classifier $\{ \psi, w_t, b_t \}$. For task-incremental learning setup, we use a shared linearized feature extraction network $g_{lin}(x; \psi)$ along with a linear classifier assigned to each task. We regularize only the feature extractor $\psi$ as each task-specific classifier does not interfere with each other.
For class-incremental learning, we train the feature extractor and append output units to the classifier at the beginning of each task.

Finally, notice that when combined with linear model, this change makes the objective function fully quadratic with respect to the parameters. This enables us to accurately represent the objective function while allowing the model to be highly accurate and expressive. The difference that these changes bring is that because the model is linear with respect to its parameters and the loss function is mean-squared error, the objective becomes quadratic with respect to the parameters. Most notably, it follows that the quadratic parameter penalty is the optimal strategy to represent the source task objective for continual learning, \ie,
\begin{align}\begin{split}
    \ell(f(x;\theta,t),y) &= \frac{1}{2} ||\alpha\phi(y) - f(x;\theta,t)||^2\\
    &= \frac{1}{2} (\theta - \theta_t)^\top H_\theta (\theta - \theta_t) + C ,
\end{split}\end{align}
where $\theta_t$ is the trained parameters after the $t$-th task. We can apply any curvature approximation algorithm to efficiently store $H_\theta$ in memory. \eg, TK-FAC~\cite{tkfac21}. The final objective function for continual learning is,
\begin{align}
    \mathcal{L}_{data} &= \E_{(x,y)\sim\mathcal{D}_t}[\ell(f(x;\theta,t),y)] \\
    \mathcal{L}_{reg} &= \frac{1}{2} (\theta - \theta_{t-1})^\top H (\theta - \theta_{t-1}) \\
    \mathcal{L} &= \frac{1}{t} \mathcal{L}_{data} + \frac{t-1}{t} \mathcal{L}_{reg},
\end{align}
where the target task objective and source task objective are weighted by $1:(t-1)$ for balancing, and $H$ corresponds to the Hessian of the source task objectives.

\subsection{Classifier regularization for class-incremental problem} 
\label{subsec:dlcft-class-il}

\begin{figure}[t]
\centering
\includegraphics[width=0.7\textwidth]{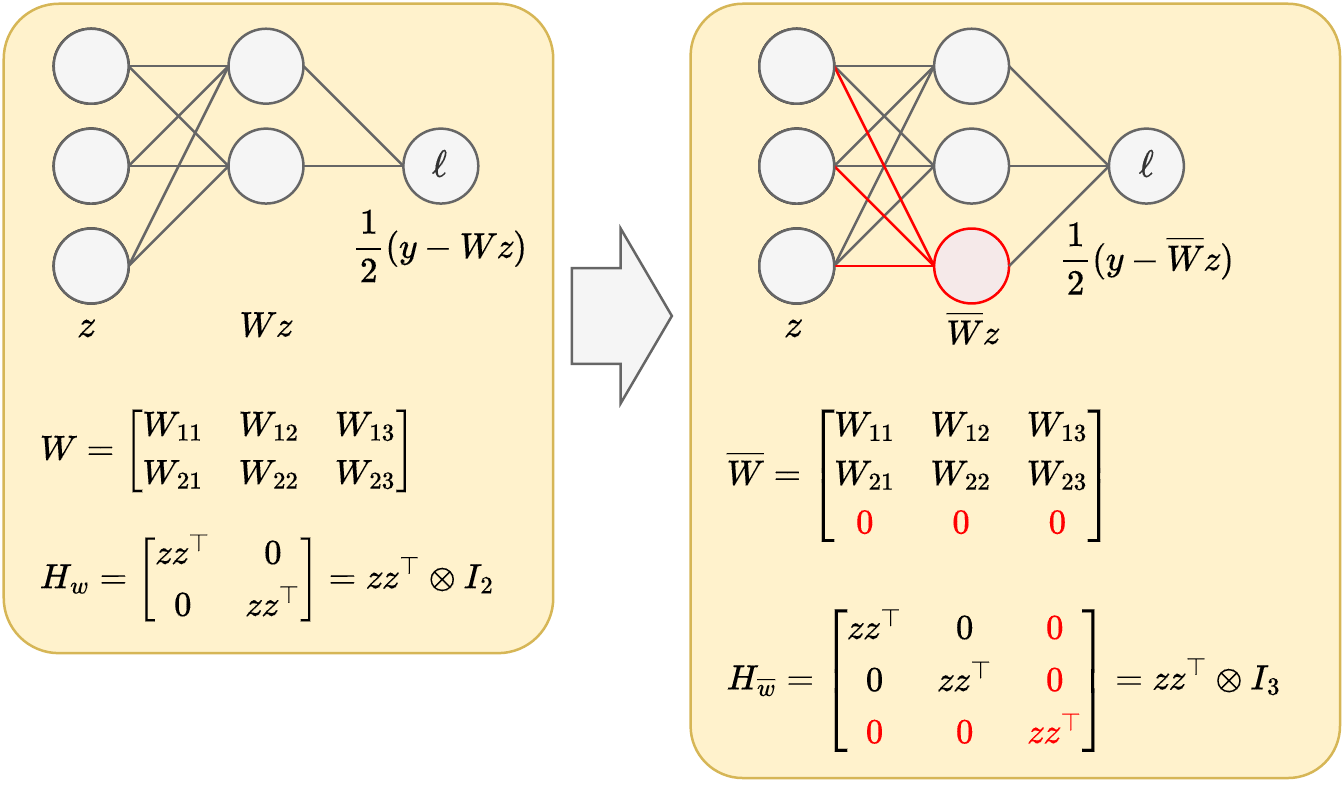}
\caption{\textbf{Curvature for classifier in class-incremental learning.} An example of a linear classification layer with two original output units and an appended unit. The appended output unit and its zero-initialized weights are outlined in red. Bias is omitted for simplicity. The corresponding updates to the weight matrix and the block diagonal of the Hessian matrix are marked in red.}
\label{fig:class-incremental}
\end{figure}

In this section, we describe how the proposed method is extended to class-incremental learning problem. To the best of our knowledge, this is the first work that shows how parameter regularization can be reasonably applied to class-incremental learning problem. 

In class-incremental learning, each task requires the model to learn a set of novel classes while the evaluation is done jointly over the current and previous tasks. Unlike task-incremental setup, task oracle is not provided at test time. At the beginning of each task, a set of output units that corresponds to the new classes is added to the classification layer. A key challenge to the problem is to apply correct regularization to the classification layer to prevent predictions from being biased towards more recent tasks.
Memory-based method achieves this by replaying samples from a buffer. On the other hand, parameter regularization methods require the curvature matrix of the source task loss to be defined for the parameters of the new output units. Here, we show that the curvature matrix is obtained without looking at the previous task data, but only from the existing curvature matrix. 
The key idea is that this is equivalent to the case where all the weights and biases of the unseen unit is set to zero. Note that this is only possible when MSE loss is used, whereas the new weights and biases diverge to infinity when SCE loss is used. Moreover, the curvature vanishes to zero when the weights and biases diverge.

Let us consider a linear classification layer with increasing output units for class-incremental setup. Adding a set of classes amounts to adding a set of corresponding output units to the weight matrix $W$. Let $\overline{W}$ be the augmented weight matrix with the added output units, and $y$ be the target. Then, for the loss function
\begin{align}
    \mathcal{L}\left( \overline{W} \right) = \E{\left[ \frac{1}{2} ||y - \overline{W}z||^2 \right]} ,
\end{align}
the second derivative with respect to $\overline{w}:=\text{Vec}\left(\overline{W}\right)$ is
\begin{align}\begin{split}
    \frac{\partial^2 \mathcal{L}}{\partial \overline{w} \partial \overline{w}^\top}  = \frac{\partial}{\partial \overline{w}} \E [(y - \overline{W}z) z^\top] 
     = \E{[ z z^\top \otimes I ]} = \E{[z z^\top]} \otimes I .
\end{split}\end{align}
Because $\E{[z z^\top]}$ has been already obtained through K-FAC regularization, we do not need additional computation or data to compute the second derivative of the appended weight.

We additionally employ a small buffer memory $\mathcal{M}$ to replay samples of previous tasks. In previous works, combining replay with parameter regularization has not been a common practice due to underwhelming performance of curvature-based regularization. However in the proposed method, the only source of error is the approximation of the curvature matrix. Whereas in replay methods, the source of error is the subsampling of source task dataset. Therefore, we can combine the proposed parameter regularization with an additional replay loss to complement for the approximation error. The final regularization objective is,
\begin{align}
    \mathcal{L}_{reg} = \frac{\lambda}{2} \cdot (\theta - \theta_{t-1})^\top H (\theta - \theta_{t-1}) + (1-\lambda)\cdot\E_{\mathcal{M}}[ \ell(f(x;\theta,t),y) ].
\end{align}

\section{Experiments}
\label{sec:experiments}

\subsection{Evaluation methods and implementation details}
\noindent \textbf{Evaluation settings.} We evaluate our method on three types of incremental learning (IL) problems: data-IL, task-IL, and class-IL.

\noindent \textbf{Models.} We use ResNet-18 \cite{resnet} architecture for all benchmarks. For the linearized ResNet-18, we followed \cite{LQF21} and replace all ReLU nonlinearities with LeakyReLU~\cite{leakyrelu}. We also followed \cite{GaF20} and folded the batch norm parameters into the convolution layers. 

\noindent \textbf{Pre-training.} For the experiments using CIFAR-100 dataset~\cite{CIFAR}, we use ImageNet32~\cite{chrabaszcz2017downsampled} which consists of 32$\times$32 downsampled images of ImageNet-1k dataset~\cite{ILSVRC15, chrabaszcz2017downsampled}. At pre-training phase, we train the model for 100 epochs using SGD optimizer with learning rate $\eta=10^{-1}$, batch size = $256$, weight decay = $10^{-5}$. We use the cosine annealing \cite{loshchilov2016sgdr} learning rate schedule. For the experiments using MIT-67 dataset~\cite{MIT67}, we use the pretrained ResNet-18 model downloaded using TorchVision\footnote{https://github.com/pytorch/vision}, which trained on ImageNet-1k. To obtain the model with LeakyReLU nonlinearties, we replicate the scheme from \cite{LQF21} and fine-tune the downloaded model on ImageNet-1k for an additional epoch using SGD with learning rate $\eta=10^{-4}$. 

\noindent \textbf{Datasets.} For data-IL setting, we used Seq-CIFAR-100 dataset~\cite{CIFAR, icarl17} split into 10 and 100 tasks, each task having 5000 and 500 samples, respectively. Additionally, we used Seq-MIT-67 for large-resolution dataset, which is MIT-67 dataset~\cite{MIT67} split into 4 tasks. For task-IL and class-IL settings, we used Seq-CIFAR-100 with 10 tasks each containing a disjoint set of classes. 

\noindent \textbf{Curvature approximation.} For approximation of the Hessian matrix, we use K-FAC~\cite{KFAC2015, kfc16} and TK-FAC~\cite{tkfac21}.

\noindent \textbf{Data augmentation.} For Seq-CIFAR-100, we apply random crop with 4 pixels of zero padding, followed by random horizontal flip. For Seq-MIT-67, we first apply resizing to 256$\times$256 then apply random crop to 224$\times$224. At test time, we resize to 256$\times$256 and apply center crop to 224$\times$224.

\noindent \textbf{Training scheme and hyperparameters.} For training nonlinear models, we used softmax cross-entropy loss and SGD optimizer with initial learning rate $10^{-3}$ and momentum 0.9. For training linearized models, we used MSE loss and Adam optimizer~\cite{kingma2014adam} with initial learning rate $10^{-4}$ and $(\beta_1, \beta_2) = (0.9, 0.999)$. In data-IL and task-IL experiments, we enable batch normalization~\cite{ioffe2015batch} at the first task only. For the loss used in class-IL, we set $\lambda = 1/2$. 

\noindent \textbf{Other implementation details.} We used Nvidia RTX 3090 GPUs and PyTorch to conduct experiments. To add a pair of Kronecker-factored curvature matrices that each correspond to the source task and the target task, we take the weighted sum the factorized matrices by $(t-1):t$ for each factors. For all experiments, we used weight decay rate of $10^{-5}$. For methods that uses buffer memory, we set the size of the buffer to 500 samples. We used reservoir sampling strategy to update the buffer.

\subsection{Incremental learning benchmarks}

\subsubsection{Data-incremental learning} 
We benchmark our method on data-IL setup in three different settings. Firstly, we tested on ten splits of Seq-CIFAR-100 training set to simulate a short sequence of data streams. Secondly, we tested on a hundred splits of Seq-CIFAR-100 training set to evaluate how the proposed method scales to a very long sequence. Finally, we tested on four splits of Seq-MIT-67 training to evaluate the methods on a high-resolution images.

\begin{figure}[t]
\centering
\begin{subfigure}{.45\textwidth}
    \centering
    \includegraphics[width=\linewidth]{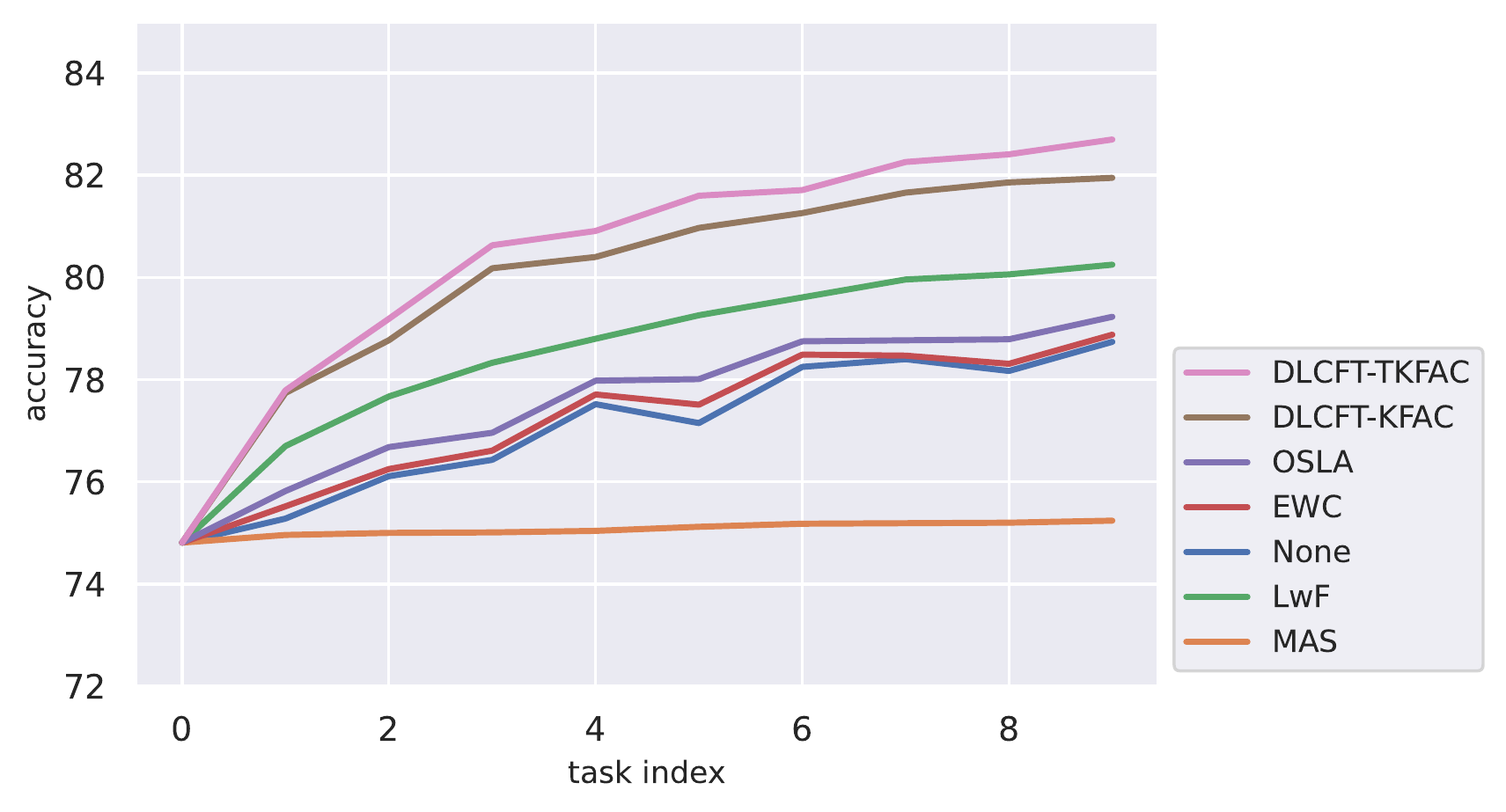}
    \caption{Seq-CIFAR-100 (10 tasks)}
\end{subfigure}
\begin{subfigure}{.45\textwidth}
    \centering
    \includegraphics[width=\linewidth]{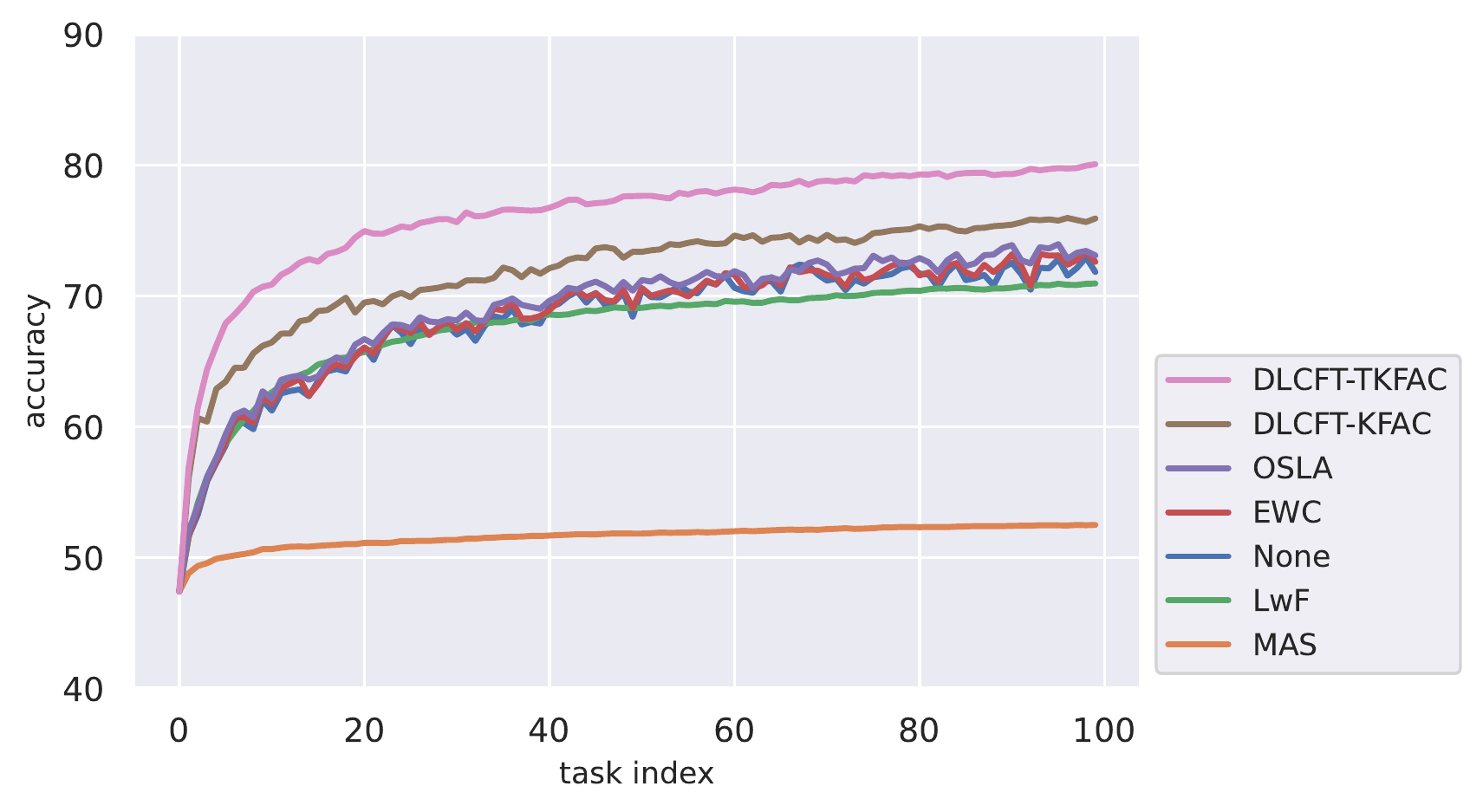}
    \caption{Seq-CIFAR-100 (100 tasks)}
\end{subfigure}
\caption{\textbf{Data-IL evaluated after each task.} Data-incremental learning using Seq-CIFAR-100. Here, we show the progress of the performance increase by evaluating after each task. Accuracy is evaluated on CIFAR-100 test set.}
\label{fig:data-inc-hundredfold}
\end{figure}

\cref{tab:experiment-results-data-il} shows performance in data-IL measured by final accuracy. We observed that Memory Aware Synapses (MAS)~\cite{mas18} fails to learn incrementally as it does not estimate accurate weight importance. On the other hand, our method performs much better than the baselines and achieves comparable performance to the joint training as it captures the curvature of the loss of the previous tasks accurately.
Additionally, in \cref{fig:data-inc-hundredfold}, we show the performance trends plotted against increasing data-IL tasks. Notably, we observed that the proposed method significantly outperforms the baselines on the long sequence length setup. 

\subsubsection{Task-/class-incremental learning} 
\cref{tab:experiment-results-task-class-il} shows performance comparison on task-incremental setup. The benchmark consists of 10 tasks with disjoint class categories obtained from CIFAR-100, each consisting of 10 classes. We observe that our method performs better than the baselines.

\newcommand{\DLCLKFAC}{$\text{DLCFT}^\dagger$}
\newcommand{\DLCLTKFAC}{$\text{DLCFT}^\ddagger$}

\begin{table}[t]
\caption{\textbf{Experiment results for data-incremental learning.} `None' indicates no regularization applied during incremental learning. The experiments are averaged over three runs.
$\dagger$ denotes that K-FAC~\cite{KFAC2015, kfc16} is used for curvature approximation and $\ddagger$ denotes TK-FAC~\cite{tkfac21} is used.
}
\medskip
\label{tab:experiment-results-data-il}
\centering
\begin{tabular}{c|cc|c}
\toprule
\textbf{Dataset} & \multicolumn{2}{c|}{\textbf{Seq-CIFAR-100}} & \textbf{Seq-MIT-67} \\
Sequence length & 10 & 100 & 4 \\
\midrule
None & 78.74 & 71.83 & 63.48 \\
LwF~\cite{LWF18} & 80.25 & 70.95 & 67.21 \\
EWC~\cite{EWC2017} & 78.88 & 72.61 & 63.68 \\
MAS~\cite{mas18} & 75.24 & 52.50 & 62.49 \\
OSLA~\cite{OSLA2018} & 79.23 & 73.10 & 64.08 \\
\DLCLKFAC (Ours) & \textbf{81.95} & \textbf{75.92} & \textbf{70.55} \\
\DLCLTKFAC (Ours) & \textbf{82.70} & \textbf{80.07} & \textbf{70.52} \\
Joint & \multicolumn{2}{c|}{83.57} & 74.40 \\
\bottomrule
\end{tabular}
\end{table}

\begin{table}[t]
\caption{\textbf{Experiment results for task-/class-incremental learning}. Experiments are conducted using Seq-CIFAR-100 with 10 tasks.
TK-FAC~\cite{tkfac21} is used for our method.
}
\medskip
\label{tab:experiment-results-task-class-il}
\centering
\begin{tabular}{cc|cc}
\toprule
Buffer size & Method & Task-IL & Class-IL \\
\midrule
\multirow{4}{*}{0} & LwF~\cite{LWF18} & 92.16 & - \\
 & EWC~\cite{EWC2017} & 77.44 & - \\
 & OSLA~\cite{OSLA2018} & 81.03 & - \\
 & DLCFT (Ours) & \textbf{95.79} & - \\
\midrule
\multirow{4}{*}{500} & ER~\cite{riemer2018learning} & 79.14 & 43.52 \\
 & DER~\cite{buzzega2020dark} & 91.47 & 58.07 \\
 & DER++~\cite{buzzega2020dark} & 91.56 & 53.29 \\
 & DLCFT (Ours) & - & \textbf{59.98} \\
\bottomrule
\end{tabular}
\end{table}

\subsection{Evaluation of incrementally learned representations}
\begin{figure}[t]
\centering
\begin{subfigure}{.45\textwidth}
    \centering
    \includegraphics[width=.8\linewidth]{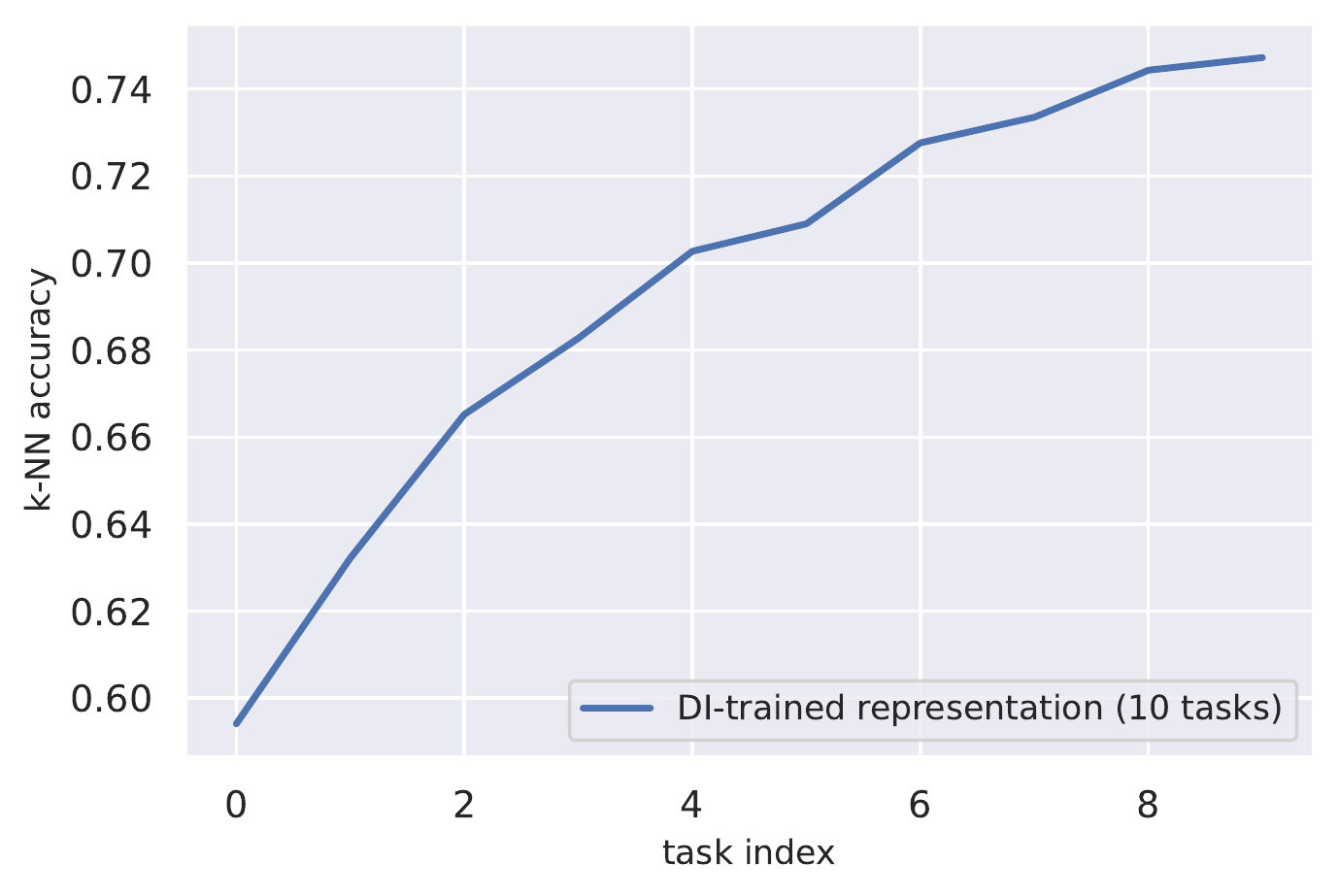} 
    \caption{Seq-CIFAR-100 (10 tasks)}
\end{subfigure}
\begin{subfigure}{.45\textwidth}
    \centering
    \includegraphics[width=.8\linewidth]{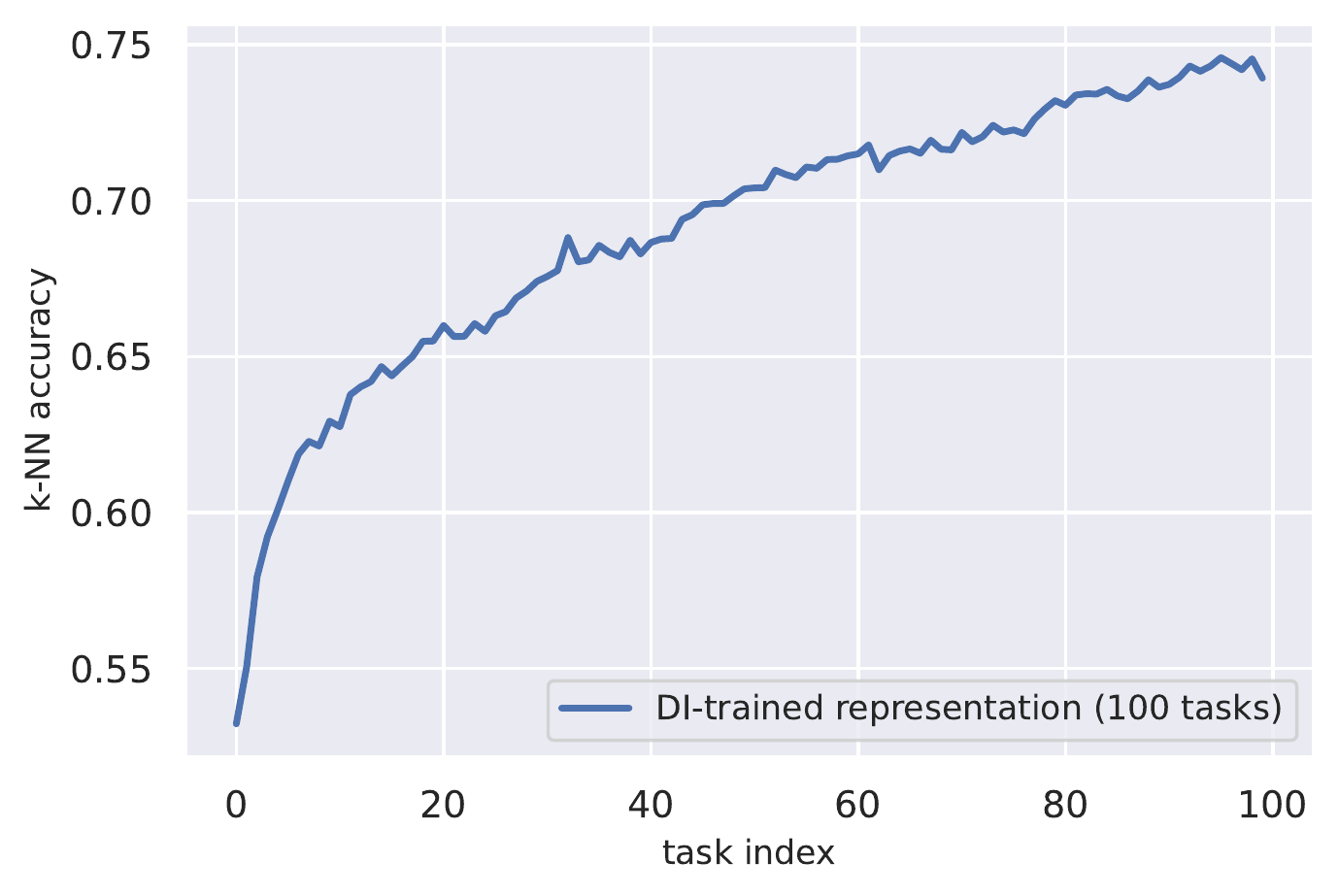}
    \caption{Seq-CIFAR-100 (100 tasks)}
\end{subfigure}
\caption{\textbf{K-NN probing of the incrementally fine-tuned models.} The quality of the representation is evaluated using K-NN classification accuracy on test set. Evaluation is performed after training each task in data-IL. At each evaluation, all training samples were used to perform inference.}
\label{fig:data-inc-knn}
\end{figure}
Because we use pre-trained networks that has already learned transferable representation from a large labeled dataset, it is possible that the performance of the method is more attributed to the linear classification layer, rather than adjusting the features through the sequential tasks. To verify that our continual fine-tuning method does learn better representations through incremental learning, we use K-NN classifier to evaluate the quality of the representations learned through the continual fine-tuning process. The results are show in \cref{fig:data-inc-knn}. The plot shows the discriminative performance of the fine-tuned feature by decoupling the linear classifier from the evaluation. We observed that the K-NN accuracy consistently and monotonically increases as the model observes more tasks.

\subsection{Ablations}

\cref{table:ablations-data-inc} shows the ablation study of the components of the proposed method performed on the data-incremental and class-incremental setting. The result shows that jointly applying the linearization and MSE loss can significantly increase performance and mitigate forgetting. Note that between nonlinear networks trained with softmax cross-entropy, employing more accurate curvature approximation brings marginal performance difference. However, between linearized networks trained with MSE loss, adopting better curvature approximation brings significant performance gain. This is because in the former case, the effectiveness of better curvature is minimal due to the vanishing curvature and higher-order error problems.

\begin{table}[t]
\caption{\textbf{Ablations.} Experiments are conducted on data- and class-IL benchmarks using Seq-CIFAR-100 dataset. `Linear' indicates linearized model, and `SCE' indicates softmax cross-entropy loss.}
\label{table:ablations-data-inc}
\medskip
\centering
\begin{tabular}{ccc|cc|c}
\toprule
 &  &  & \multicolumn{2}{c|}{\textbf{Data-IL}} & \textbf{Class-IL} \\
Curvature & Linear & Loss & 10 tasks & 100 tasks & 10 tasks \\
\midrule
\multirow{2}{*}{K-FAC~\cite{KFAC2015}} & \xmark & SCE & 79.23 & 73.51 & 44.93\\
 & \cmark & MSE & \textbf{81.95} & \textbf{75.92} & \textbf{59.86}\\
\midrule
\multirow{2}{*}{TK-FAC~\cite{tkfac21}} & \xmark & SCE & 79.58 & 73.18 & 44.66\\
 & \cmark & MSE & \textbf{82.70} & \textbf{80.07} & \textbf{59.98}\\
\bottomrule
\end{tabular}
\end{table}

\section{Conclusion}
\label{sec:conclusion}

In this paper, we have explored \emph{continual fine-tuning}, which is a practical framework for incremental learning of deep neural networks. For this, we propose Deep Linear Continual Fine-tuning, which is a simple and effective continual learning algorithm using a pre-trained neural network. We provided theoretical reasons on why existing Hessian-based parameter regularization performs poorly with neural networks trained using softmax cross-entropy loss. We showed that a combination of model linearization technique and mean-squared error loss function allows the parameter regularization methods to closely match the optimal continual learning policy. We provided a principled approach to applying parameter regularization in class-incremental learning scenario, and showed that our method outperforms other baselines on data-/task-/class-incremental settings. Moreover, we show that our method can effectively accumulate knowledge over very long data-incremental tasks sequences. 

\subsubsection*{Acknowledgements} 
This work was supported by Institute of Information \& communications Technology Planning \& Evaluation (IITP) grant funded by the Korea government(MSIT) (No. 2022-0-00951, Development of Uncertainty-Aware Agents Learning by Asking Questions).

%
%
\bibliographystyle{splncs04}
\bibliography{egbib}

\clearpage

\appendix

\newcommand{\aux}{\partial_r x_{l}}
\newcommand{\partition}{\sum_{m}{e^{x_m}}}

\section{Second derivatives of softmax cross-entropy loss}
\label{supp:sce-loss-hessian}

Here, we show the vanishing behavior of the Hessian of the softmax cross-entropy loss. Let us consider a probabilistic classification model $p(y|x)$ where $x \in \mathbb{R}^M$ is the input to the softmax layer. The model is defined as,
\begin{align}
    p(y=k|x) = \frac{e^{x_k}}{\sum_{m}{e^{x_m}}}.
\end{align}
Assuming that the target label is $t$, the cross-entropy loss is,
\begin{align}
    \mathcal{L}(x) = - \log{p(y=t|x)} = -x_t + \log{\sum_{m}{e^{x_m}}}.
\end{align}
Then, the second derivatives of the softmax cross-entropy is,
\begin{align}
    \frac{\partial^2 \mathcal{L}(x)}{\partial x_i \partial x_j} & = \frac{\partial^2}{\partial x_i \partial x_j} \left( -x_t + \log{\sum_{m}{e^{x_m}}} \right) \\
    &= \frac{\partial^2}{\partial x_i \partial x_j} \left(\log{\sum_{m}{e^{x_m}}} \right) \\
    &= \frac{\partial}{\partial x_j} \left(\frac{e^{x_i}}{\sum_{m}{e^{x_m}}} \right).
\end{align}
The diagonal entries of the Hessian where $j=i$ is,
\begin{align}
    \frac{\partial^2 \mathcal{L}(x)}{\partial x_i^2} = \frac{e^{x_i}\left(\partition - e^{x_i}\right)}{\left(\partition\right)^2}
    = \frac{\frac{e^{x_i}}{e^{x_t}}\left(1+\sum_{m\neq t}{\frac{e^{x_m}}{e^{x_t}}} - \frac{e^{x_i}}{e^{x_t}}\right)}{\left(1+\sum_{m\neq t}{\frac{e^{x_m}}{e^{x_t}}}\right)^2}.
    \label{eq:hessian-diagonal}
\end{align}
And the off-diagonal entries where $j\neq i$ is,
\begin{align}
    \frac{\partial^2 \mathcal{L}(x)}{\partial x_i \partial x_j} = \frac{- e^{x_i}e^{x_j}}{\left(\partition\right)^2}
    = \frac{- \frac{e^{x_i}}{e^{x_t}}\frac{e^{x_j}}{e^{x_t}}}{\left(1+\sum_{m\neq t}{\frac{e^{x_m}}{e^{x_t}}}\right)^2}.
    \label{eq:hessian-offdiagonal}
\end{align}
Now, let us consider the limiting case when the model converges to the target, $p(y=t|x) \rightarrow 1$, \ie, if and only if $e^{x_m - x_t} \rightarrow 0$ for all $m\neq t$. Then, it can be seen from \cref{eq:hessian-diagonal} and \cref{eq:hessian-offdiagonal} that all the entries of the Hessian matrix converges to zero.
\clearpage

\section{Implementation of linearized neural network}
\label{appendix:forward-pass-implementation}

Here, we provide implementation details for linearized nerual network. We consider a network with pre-trained parameter $\theta_0$. We denote the neural network as $f(x; \theta_0)$. Then we apply first-order Taylor approximation with respect to the parameters to linearize the network around $\theta_0$ as,
\begin{equation}
    f(x; \theta_0 + \Delta\theta) \simeq f(x; \theta_0) + D_\theta f(x; \theta_0) \cdot \Delta\theta,
\end{equation}
where $D_\theta f(x; \theta_0)$ is the Jacobian of the network evaluated at $(x, \theta_0)$. For most neural networks, the Jacobian matrix is prohibitively expensive to compute and store due to the size of parameter dimension. 
To compute the forward pass, we use the modified forward pass method proposed in \cite{GaF20} based on forward-mode automatic differentiation algorithm, which efficiently computes Jacobian-vector products (JVP). Unlike backpropagation, the algorithm does not require extra memory footprint to compute the derivatives.

\subsection{Augmented forward propagation}

We implemented the modified forward pass by subclassing the layers in PyTorch~\cite{NEURIPS2019_9015} library.\footnote{\href{https://github.com/pytorch/pytorch}{https://github.com/pytorch/pytorch}} The inputs and outputs of the custom layers are a tuple of hidden state and augmented state, such that the forward pass jointly computes the activations and the JVP.

\cref{table:augmented-forward-formula} shows the formulas for the custom layer implementation. We use the same notation as \cite{GaF20} and denote the augmented state for JVP as $\aux$. The JVP for full neural network is computed by feeding a zero-initialized vector having the same shape of the input as the incoming augmented input to the first layer. Finally, upon the completion of the forward pass, the computed hidden state and augmented state are summed to obtain the output of the linearized neural network.

\begin{table}[H]
\caption{Formulas for the augmented forward pass implementation. $\mathbbm{1}\{x>0\}$ indicates a mask vector that has value 1 where the condition is satisfied and 0 otherwise. Max pooling layer is decomposed into MaxPoolIndices$(\cdot)$ and Gather$(\cdot)$ operations which gives the indices of the pooled values and aggregates the values using the indices.}
\label{table:augmented-forward-formula}
\centering
\medskip
\resizebox{\columnwidth}{!}{
\begin{tabular}{c|c|c}
\toprule
Layer Type & Hidden State & Augmented State \\
\midrule
Identity & $x_l$ & $\aux$ \\
Linear, Conv & $f(x_l; W; b)$ & $f(x_l; \Delta W, \Delta b) + f(\aux; W, 0)$ \\
ReLU & $x_l \odot \mathbbm{1}\{x_l>0\}$ & $\aux \odot \mathbbm{1}\{x_l>0\}$ \\
LeakyReLU & $x_l \odot \mathbbm{1}\{x_l>0\} + \alpha x_l \odot \mathbbm{1}\{x_l\leq0\}$ & $\aux \odot \mathbbm{1}\{x_l>0\} + \alpha \aux \odot \mathbbm{1}\{x_l\leq0\}$ \\
MaxPool & $\text{Gather}(x_l, \text{MaxPoolIndices}(x_l))$ & $\text{Gather}(\aux, \text{MaxPoolIndices}(x_l))$ \\
AveragePool & AvgPool$(x_l)$ & AvgPool$(\aux)$ \\
\bottomrule
\end{tabular}
}
\end{table}

\section{Additional experiments on vanishing curvature}

Here, we show the curvature behavior during training using CIFAR-100 dataset.

\begin{figure}[h]
  \centering
    \begin{subfigure}{.40\linewidth}
        \centering
        \includegraphics[width=0.9\linewidth]{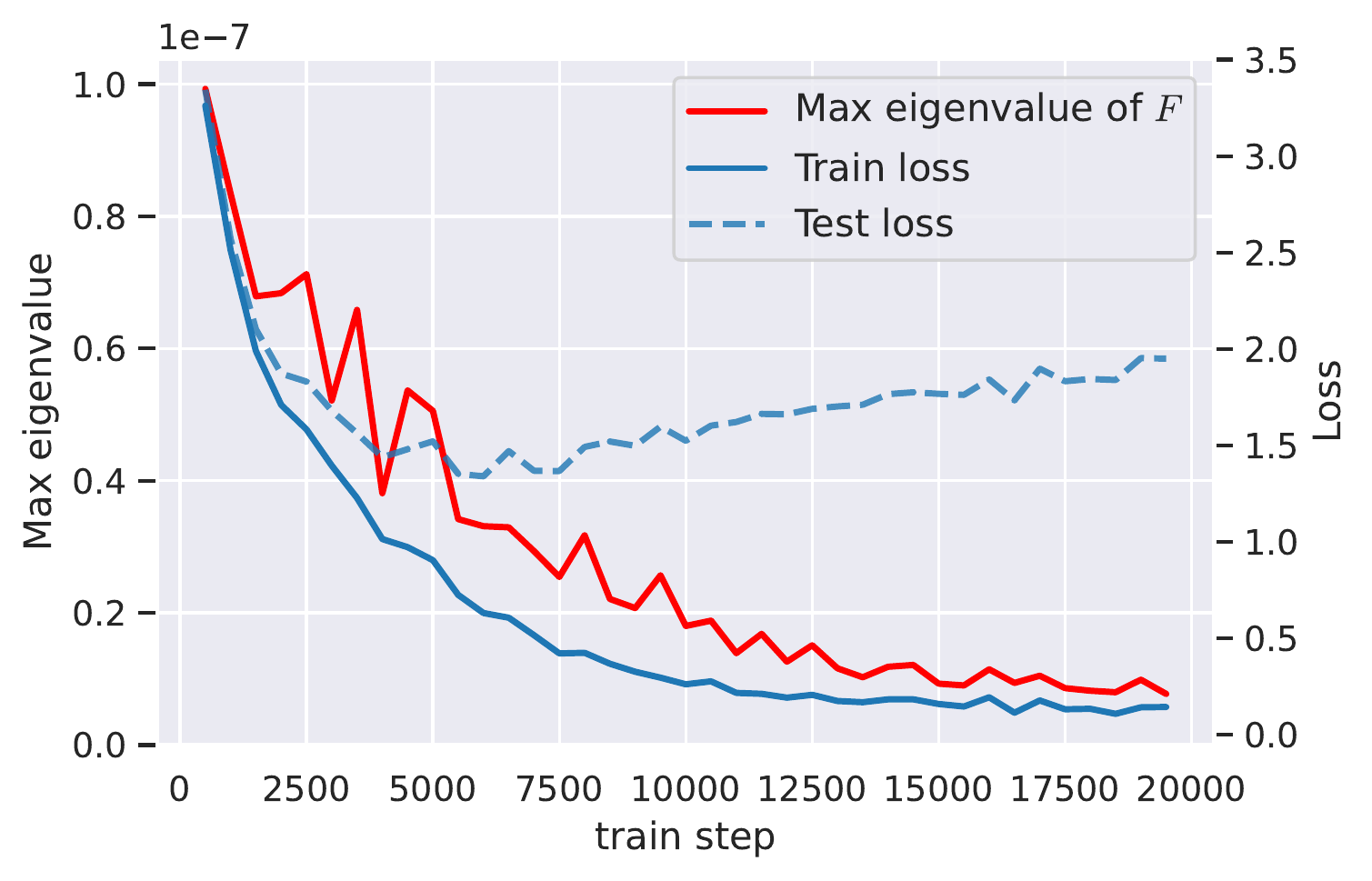}
        \caption{Curvature of SCE loss}
    \end{subfigure}
    \begin{subfigure}{.40\linewidth}
        \centering
        \includegraphics[width=0.9\linewidth]{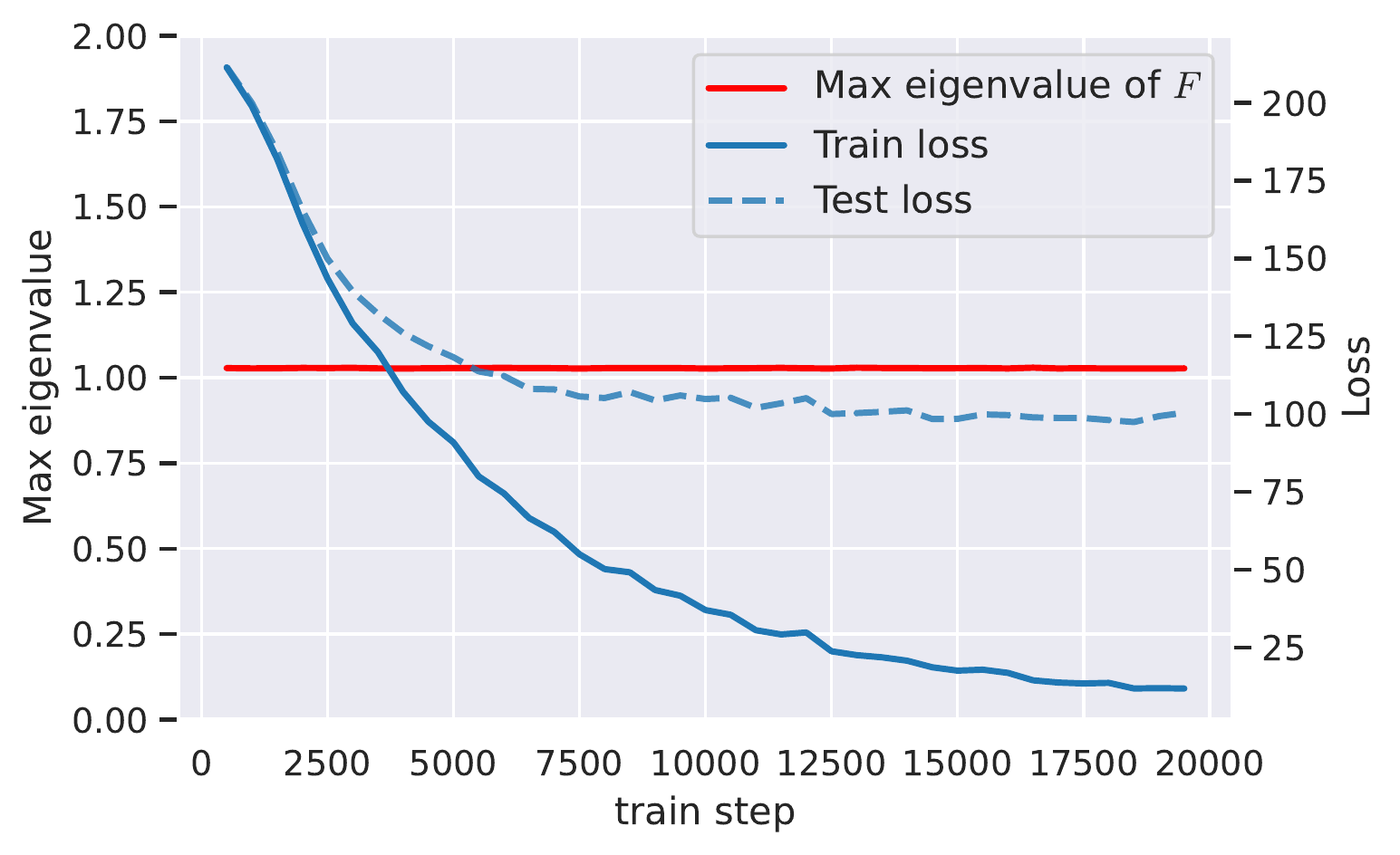}
        \caption{Curvature of MSE loss}
    \end{subfigure}
  \caption{Maximum eigenvalue of the FIM during training using CIFAR-100 dataset. 
  }
  \label{fig:eigenvalues}
\end{figure}

\section{Comparison of curvature methods on data-IL}

Here, we demonstrate the performance impact when combined with different curvature approximations. We highlight that the proposed method leads to performance gain in all types.

\begin{table}[h]
\caption{Performance comparison of curvature methods on data-IL using Seq-CIFAR-100 with 10 tasks.}
\label{table:ablations-curvature}
\medskip
\centering
\begin{tabular}{c|cc}
\toprule
Curvature & Nonlinear+SCE & Linear+MSE \\
\midrule
EWC & 78.88 & 82.17 \\
K-FAC & 79.23 & 81.95 \\
TK-FAC & - & 82.70 \\
\bottomrule
\end{tabular}
\end{table}
\clearpage

\section{Backward transfer evaluations on task-IL}

Here, we provide task-IL performance measured in Backward transfer (BWT) metric. Note that our method is on par with LwF and significantly outperforms other parameter regularization methods. Moreover, our method outperforms LwF in average accuracy (ACC) metric.

\begin{table}[h]
\caption{Average accuracy (ACC) and backward transfer (BWT) on task-IL using Seq-CIFAR-100 with 10 tasks.}
\label{table:backward-transfer}
\medskip
\centering
\begin{tabular}{c|cc}
\toprule
Method & ACC & BWT \\
\midrule
LWF & 92.16 & 0.04 \\
EWC & 77.44 & -21.63 \\
OSLA & 81.03 & -18.21 \\
DLCFT (Ours) & 95.79 & -0.58 \\
\bottomrule
\end{tabular}
\end{table}

\end{document}